\documentclass{article}

\usepackage{arxiv}

\usepackage[utf8]{inputenc} % allow utf-8 input
\usepackage[T1]{fontenc}    % use 8-bit T1 fonts
\usepackage{hyperref}       % hyperlinks
\usepackage{url}            % simple URL typesetting
\usepackage{booktabs}       % professional-quality tables
\usepackage{amsfonts}       % blackboard math symbols
\usepackage{nicefrac}       % compact symbols for 1/2, etc.
\usepackage{microtype}      % microtypography
\usepackage{lipsum}
\usepackage{epsfig}
\usepackage{graphicx}
\usepackage{amsmath}
\usepackage{amssymb}
\usepackage{subfig}
\usepackage{algorithm}
\usepackage[noend]{algpseudocode}

\title{RocNet: Recursive Octree Network for Efficient 3D Deep Representation}

\author{
  Juncheng Liu \\%\thanks{Use footnote for providing further
  Department of Computer Science\\
  University of Otago\\
  Dunedin, New Zealand \\
  \texttt{juncheng.liu@otago.ac.nz} \\
   \And
 Steven Mills \\
  Department of Computer Science\\
  University of Otago\\
  Dunedin, New Zealand \\
  \texttt{steven@cs.otago.ac.nz} \\
   \AND
 Brendan McCane \\
  Department of Computer Science\\
  University of Otago\\
  Dunedin, New Zealand \\
  \texttt{mccane@cs.otago.ac.nz} \\
}

\begin{document}
\maketitle

\begin{abstract}
    We introduce a deep recursive octree network for the compression of 3D voxel data. Our network compresses a voxel grid of any size down to a very small latent space in an autoencoder-like network. We show results for compressing $32^3$, $64^3$ and $128^3$ grids down to just 80 floats in the latent space. 
    We demonstrate the effectiveness and efficiency of our proposed method on several publicly available datasets with three experiments: 3D shape classification, 3D shape reconstruction, and shape generation. Experimental results show that our algorithm maintains accuracy while consuming less memory with shorter training times compared to existing methods, especially in 3D reconstruction tasks.

\end{abstract}

%%%%%%%%% BODY TEXT
\section{Introduction}

While neural networks achieve excellent performance in various tasks in 2D vision, how to effectively process 3D data by neural networks has recently attracted more attention in both the computer vision and computer graphics communities. Unlike the unified representation of 2D images as pixels, there are many formats widely used for 3D data such as point clouds, volumetric voxels and surface meshes. Each of these formats has its advantages and disadvantages. Point clouds represent 3D data by unstructured points without imposing topological constraints. However, they bring difficulties in  subsequent applications such as rendering and space navigation. Surface meshes are most widely used in graphics due to efficient rendering and modeling flexibility, yet their inherent graph structure makes it difficult, though not impossible, to process via neural networks. 3D volumetric voxel grids are the straightforward generalization of 2D pixels and are a natural representation of 3D space. But voxel grids are very memory intensive and memory requirements grow very fast as the size of the grid increases, making them difficult to apply to fine resolutions or large spaces.

\begin{figure}[t]
\begin{center}
\includegraphics[width=0.5\linewidth]{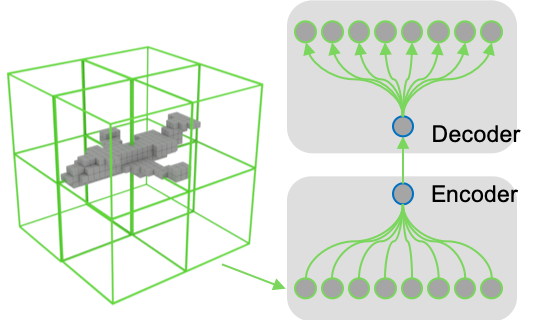}
\end{center}
   \caption{RocNet autoencoder structure. Octants are merged recursively in a bottom-up manner until the root node is encoded. Decoder recursively decodes nodes until leaves are recovered.}
\label{fig:teaser}
\end{figure}

In this paper we introduce RocNet: a recursive 3D autoencoder network that mimics the structure of an octree.
As shown in Figure~\ref{fig:teaser}, a voxel grid is first converted to a standard octree structure until each node is homogeneous or a maximum depth is reached. RocNet recursively merges octants starting from the octree leaves, until the root is processed. In this way the whole octree can be represented by a compact 1D feature vector. Since there are a large number of empty leaves in an octree, our algorithm saves large amount of memory and computational time. Compared to traditional 3D convolutional networks, 3D convolution is only performed within octants in our recursive network hence is less computationally intensive. Memory savings arise from the octree structure which adapts to the geometry of the scene, allowing large empty (or otherwise homogeneous) volumes to be represented compactly.

RocNet is an end-to-end autoencoder that is a recursive, discriminative, and generative network that is primarily aimed at compressing 3D voxel structures for representation, discrimination, and generation. The main contributions of this work are:
this is the first method to explore the combination of volumetric octree and recursive networks;
near state-of-the-art compression ratios;
near state-of-the-art recognition on ModelNet40;
state-of-the-art computational and memory consumption efficiency;
state-of-the-art reconstruction accuracy on ShapeNetCorev2 database.

%------------------------------------------------------------------------

%For this citation style, keep multiple citations in numerical (not
%chronological) order, so prefer \cite{Alpher03,Alpher02,Authors12} to
%\cite{Alpher02,Alpher03,Authors12}.

%------------------------------------------------------------------------
\section{Related work}

The idea of using octrees to efficiently process sparse 3D data has been intensively explored in deep learning in recent years. These methods can be roughly classified into two groups: discriminative and generative networks. 

\textbf{Discriminative octree networks.} Discriminative networks are designed to accomplish recognition tasks such as classification and retrieval. O-CNN~\cite{wang2017cnn} uses an octree-based 3D convolutional network by using fast shuffled key and hash table search. Their method significantly reduces the computational time and storage requirements for 3D convolution by restricting computations to occupied octants only. OctNet~\cite{riegler2017octnet} uses an alternative octree-based convolutional network where a more efficient convolution is defined for octants. A major difference between RocNet and the other methods is RocNet performs convolution only within octants. Convolution is not performed across octants.

\textbf{Generative octree networks.} Generative octree networks are able to generate 3D data in an octree-like manner. Adaptive O-CNN~\cite{wang2018adaptive} implements an advanced version of O-CNN which is able to generate models in adaptive patches. HSP~\cite{hane2017hierarchical} predicts high resolution voxel grids from images by using an octree up-convolutional decoder architecture. OctNetFusion~\cite{riegler2017octnetfusion} is the generative version of OctNet, which reconstructs dense 3D data from multiple depth images by predicting the partitioning of the 3D space. Similar to our method, OGN~\cite{tatarchenko2017octree} also generates octrees, but their method is aimed at up-sampling from a coarse regular voxel grid and thus focuses on decoding only. %, typically starting with a $32^3$ grid as the compressed representation. 
In contrast, we seek to form an efficient latent space for storing grids of arbitrary size, where the latent space is not restricted to be grid-like. 

\textbf{Recursive networks.} Our method is also related to recursive networks which were originally proposed as an effective tool for grammar tree parsing in natural language. In computer graphics, GRASS~\cite{li2017grass} employed  recursive networks to implement a generative shape structure autoencoder. Their work focuses specifically on object structures and makes explicit use of object symmetries. %, and thus is considerably more complicated than RocNet. 
RocNet focuses on volumetric representation and can also be used for generic 3D representations such as rooms or buildings. 
Another successful application of recursive networks is $k$-d networks~\cite{klokov2017escape} which use a $k$-d tree to hierarchically process 3D point clouds, whereas we focus on voxel representations.

\begin{figure}
\begin{center}
\includegraphics[width=0.5\linewidth]{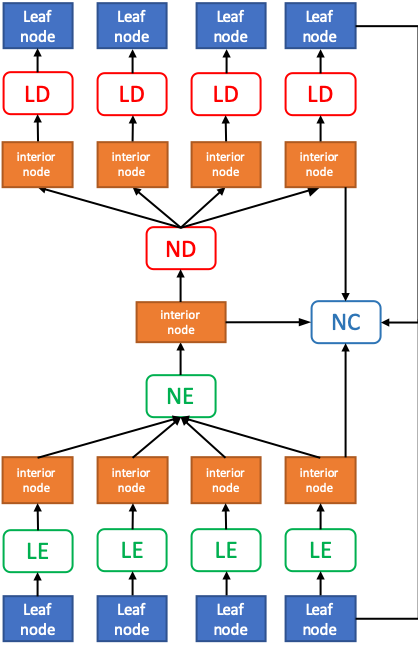}
\end{center}
   \caption{RocNet autoencoder structure. LE: leaf encoder. NE: node encoder. ND: node decoder. LD: leaf decoder. NC: node classifier.}
\label{fig:rocnet}
\end{figure}

\section{Recursive octree networks}

In this section we will describe the network architecture by which the octrees are encoded and decoded. The overall structure of the proposed network is shown in Figure~\ref{fig:rocnet}. The network has a recursive structure that recursively merges 8 child nodes to form their parent node until the root node is encoded. The insight is that the recursive feature easily fits the hierarchical nature of octrees. 

Our network consists of 7 components: leaf encoder, node encoder, tree encoder, leaf decoder, node decoder, tree decoder and node classifier.  The octree is encoded from the leaves up. The {\em leaf encoder} encodes the leaves and produces a multi-channel 3D code for each leaf. For each non-leaf node, the {\em node encoder} combines codes from each of its children until the root is reached, again producing a multi-channel 3D code for each node. Finally the {\em tree encoder} converts a multi-channel 3D code into a compact 1D feature vector. To produce an octree from a compact code, the corresponding {\em decoders} reverse the encoding process. The only difference is that we need to determine if a given node is an internal node or a leaf and the {\em node classifier} is used to make that determination. We describe the octree representation and each component in more detail in the following sections.

\begin{figure*}
\begin{center}
	\subfloat[]{
		\label{fig:octree1}
		\centering
		\includegraphics[width=0.20\linewidth]{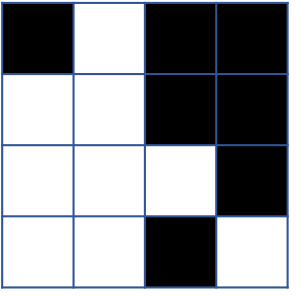}
	}
	\subfloat[]{
		\label{fig:octree2}
		\centering
		\includegraphics[width=0.10\linewidth]{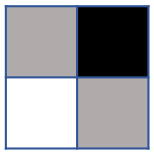}
	}
	\subfloat[]{
		\label{fig:octree3}
		\centering
		\includegraphics[width=0.06\linewidth]{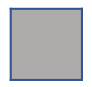}
	}
	\subfloat[]{
		\label{fig:octree4}
		\centering
		\includegraphics[width=0.20\linewidth]{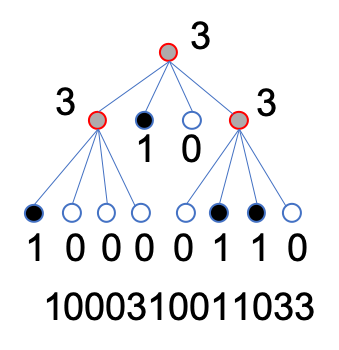}
	}
	\subfloat[]{
		\label{fig:octree5}
		\centering
		\includegraphics[width=0.13\linewidth]{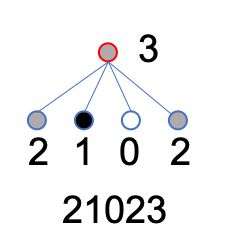}
	}
	\subfloat[]{
		\label{fig:octree6}
		\centering
		\includegraphics[width=0.2\linewidth]{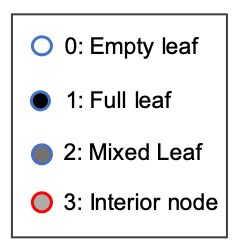}
	}
\end{center}
   \caption{Example of an octree, although a quad-tree is shown for simplicity. (a)-(c) depth 2 to 0~(root) of an occupancy octree. Black square indicates occupied octants and white square indicates empty ones. Mixed nodes are represented as grey squares. (d) octree structure of maximum depth 2 and its corresponding structure code. (e) octree structure of maximum depth 1 and its corresponding structure code. (f) 4 defined node types.} 
\label{fig:octree}
\end{figure*}

\subsection{Octree representation}

An octree decomposes a 3D space recursively until it contains homogeneous content (all voxels inside a leaf should be consistently occupied or empty) or a given maximum depth is reached. In the proposed algorithm, we adopt the octree representation as the input format of our neural network. %This brings us three main advantages: (1) it saves large amount of memory; (2) less computation; (3) an implementation of multi-resolution reconstruction. 

An example of an octree structure is presented in Figure~\ref{fig:octree}. For simplicity, we illustrate a quad-tree on a 2D grid instead of an actual octree on a 3D grid. We define 4 types of octants: empty leaf, full leaf, mixed leaf and interior node as shown in Figure~\ref{fig:octree6}. %An octant will be decomposed recursively unless it satisfies one of the three criterion: (1) its voxels are all empty. (2) its voxels are all occupied. (3) the given maximum depth is reached. 
An octree can be therefore represented as two separate parts: tree topology and its mixed leaves. Empty and full leaves do not need storage and can be directly recovered from tree topology.  

\textbf{Tree topology representations.} To encode octree topology, we use a post-order Depth-First Search (DFS) ordering. As shown in Figure~\ref{fig:octree4}, the octants of a 3 layer octree are serialized by the post-order DFS: it starts at the root node and goes as far as it can down its child nodes from left to right, then visit itself and backtracks. 

\textbf{Mixed leaf representations.} We collect all mixed leaf nodes in an octree in the order defined above, forming a 4D binary matrix. In the example shown in Figure~\ref{fig:octree5}, the maximum depth is 1 and two mixed leaf nodes are extracted. To clearly demonstrate the advantages of our proposed structure, we simply use binary occupancy encoding across all our experiments. However, other representations such as real-valued occupancy probability or the surface normal could also be easily integrated into our algorithm in this step.

\begin{figure}
\begin{center}
	\subfloat[encoder]{
		\label{fig:encoder}
		\centering
		\includegraphics[width=0.20\linewidth]{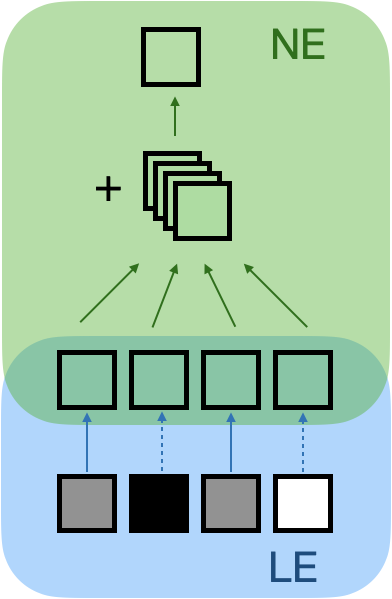}
	}
	\subfloat[decoder]{
		\label{fig:decoder}
		\centering
		\includegraphics[width=0.20\linewidth]{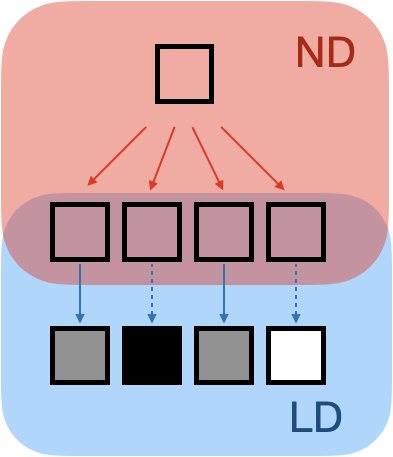}
	}
\end{center}
   \caption{Node encoder and decoder. Arrow lines are 3D convolution and batch normalization. Dotted lines are empty/full leaf skips. Quad-tree is shown here for simplicity.}
\label{fig:node}
\end{figure}

\subsection{Recursive octree encoder}
\label{sec:encoder}

Generally speaking, the encoders encode leaf nodes, interior nodes, and eventually the whole octree into features as shown in Figure~\ref{fig:encoder}. 

\textbf{Leaf encoder.} Explicit leaf nodes need to be converted into recognizable features before being fed into the recursive network. In comparison with a traditional recursive autoencoder~\cite{socher2014recursive} (RAE) that encodes nodes into 1D features, our method encodes a leaf node into a 4D feature with fixed dimensions using several 3D convolutional layers of kernel size 4 and stride 2. The number of channels of feature maps is set to 16, 32 and 64. We use the exponential linear unit~\cite{clevert2015fast} (ELU) function $(f : x \in \mathcal{R} \mapsto \max(0,x) + \min(0,e^x - 1))$ as the activation function and add a batch normalization (BN) layer after each 3D convolution layer. Our network is free of a pooling layer. Therefore the sequence of a leaf encoder is several layers of ``3DConv + BN + ELU ''. The leaf encoder therefore is a mapping: $\mathcal{B}^{ k \times k \times k} \mapsto \mathcal{R}^{64\times4\times4\times4}$, where $\mathcal{B}\in\{0,1\}$, $k$ is the leaf resolution~(we set $k=32$ for voxels larger than $32^3$ and $k=16$ for  $32^3$), $64$ is the number of output channels. Since there are a large number of empty and full leaf nodes, applying 3D convolution to them is unnecessary and time-consuming, hence we pass these leaves directly to the next stage without applying 3D convolutions. We call these ``leaf skips'' as shown in Figure~\ref{fig:encoder} and \ref{fig:decoder}.

\textbf{Node encoder.} It first lifts the number channels of child nodes from 64 to 128 by a $1\times1\times1$ sized kernel 3D convolutional layer denoted as $\phi: \mathcal{R}^{64\times4\times4\times4} \mapsto \mathcal{R}^{128\times4\times4\times4}$ and then merges the nodes into an interior parent node. The inverse of $\phi$, denoted as $\psi$, is applied to the parent node which maps number of channels back to 64. We use ``additive merging'' for all encoders of this kind, that is, every child node is first applied with a 3D convolution followed by a batch normalization layer, and then added together to form the parent node:
\begin{equation}
F_p = \psi(\sum_{i=1}^{8}{\text{elu}(\phi(F_i)))}, 
\end{equation}
$F_p: \mathcal{R}^{64\times4\times4\times4} \mapsto \mathcal{R}^{64\times4\times4\times4}.$
This encoder is applied recursively until the root node is encoded. Note that node encoders for different depth do not share weights.

\textbf{Tree encoder.} After obtaining the feature of the root node, the whole octree is encoded to a single feature map of $64\times4\times4\times4$. The tree encoder is applied to flatten the feature map into a single 1D feature vector. We achieve this by simply employing a 3D convolutional layer with kernel size 4 and stride 1. The number of channels is set to the dimensions of the feature vector. The tree encoder then is a map: $\mathcal{R}^{64\times4\times4\times4} \mapsto \mathcal{R}^{d_{out}}$. In our implementation $d_{out}$ is set to 80 across all experiments.

\subsection{Recursive octree decoder}
\label{sec:decoder}

The decoders simply reverse the above process using transposed convolutions instead of convolutions.

\textbf{Tree decoder.} This decoder converts a 1D feature back to a 4D feature by a transposed convolutional layer and a non-linear activation layer ELU. To transform feature vectors back to the above defined dimensions, we use a kernel of size 4 and stride 1.

\textbf{Node decoder.} After decoding the tree we obtain the 4D feature for the root node. We then apply the node decoder to it recursively until leaf nodes are decoded. This decoder consists of a 3D convolutional layer for the node itself and a convolutional layer for each its 8 child nodes followed by a batch normalization layer as shown in Figure~\ref{fig:decoder}.  Note that node decoders for different depth do not share weights.

\textbf{Leaf decoder.} When a node is recognized as a leaf it will be decoded by a leaf decoder which recovers the features back to explicit binary occupancy voxel grids. Note that leaf skips will also be applied in the leaf decoder for efficiency, i.e. only leaves recognized as mixed need to be decoded. This decoder consists of several stacked transposed 3D convolutional layers of kernel size 4 stride 2 and padding 1. This operation will enlarge the voxel size from 4 to $k$ which is the dimension of explicit representation we used for leaves. All layers are followed by batch normalization layers and ELU activation with the exception of the final layer which is activated by a sigmoid non-linearity without batch normalization.

\subsection{Node classifier} 
A node classifier is a 4-category classifier trained during the decoding process in order to recover the octree topology. Note that the classifier is not involved in encoding, as at that stage the type of each node is known. This classifier labels each node one of the 4 aforementioned node types. It has the same layers as tree encoder except for an additional fully-connected layer. We use cross-entropy as the loss function. The node classifier is disabled during training and is only used for prediction in the testing stage. The predicted label is used to decide whether to stop decoding~(empty/full leaf) or to decode using leaf decoder (mixed leaf) or node decoder (interior node).

\subsection{Loss function.} The loss function in our algorithm has two separate parts: the node labeling loss, $\mathcal{L}_l$ and the leaf reconstruction loss, $\mathcal{L}_r$:
\begin{equation}
\mathcal{L} = \mathcal{L}_l + \mathcal{L}_r.
\end{equation}

\textbf{Node labeling loss.} We use cross-entropy loss given by node classifier as the node labeling loss:
\begin{equation}
\mathcal{L}_l = -\sum_{i}^{4} c_i \log(s_i),
\end{equation}
where $c_i$ and $s_i$ are the ground-truth and node classifier score for 4 node types. The node label loss of an octree is the sum of label losses of all nodes.

\textbf{Reconstruction loss.} Decoded leaves are used to calculate the reconstruction loss and we use a weighted binary cross-entropy as the loss function:
\begin{equation}
\mathcal{L}_r = -\alpha t\log(o) - (1-t) \log(1-o),
\end{equation}
where $t \in \{0,1\}$ is the ground-truth occupancy value and $o \in (0,1)$ is the output of the leaf decoder for each individual voxel. The weight parameter $\alpha$ is employed because the number of empty voxels is usually much larger than occupied ones. In most cases, $\alpha$ should be larger than 1 for elimination of false negatives. In our implementation we use $\alpha=5$ across all experiments. The reconstruction loss of an octree is the sum of the reconstruction losses across all its mixed leaves.

\begin{algorithm}
\caption{RocNet node encoder and decoder}\label{ag:rocnet}
\begin{algorithmic}
\Procedure{Encoding(node)}{}
\If {\textit{node.is\_leaf()}} \Return \textit{LeafEncoder(node)}
\Else
\For{$k \gets 1$ to $8$}  
\State {$child$ $\gets$ {$child$ + \textit{Encoding(node.get\_child($k$))}}}
\EndFor
\State \Return $child$
\EndIf
\EndProcedure

\Procedure{Decoding(node)}{}
\If {\textit{node.pred\_leaf()}}  \textit{LeafDecoder(node)}
\Else
\For{$k \gets 1$ to $8$}  
\State {\textit{Decoding(node.get\_child($k$))}}
\EndFor
\EndIf
\EndProcedure
\end{algorithmic}
\end{algorithm}

\subsection{Complexity analysis}
\label{sec:complexity}
\begin{table}
\begin{center}
\begin{tabular}{|c|c|c|c|}
\hline
 & size & & size\\
\hline\hline
PointNet~\cite{garcia2016pointnet} & 80M  & RocNet-64-32 & 1.42M \\
3DShapeNets~\cite{wu20153d} & 38M & RocNet-128-32 & 1.70M \\
VoxNet~\cite{7353481} & 0.92M & RocNet-256-32 & 1.98M \\
LightNet~\cite{zhi2018toward} & 0.3M & RocNet-512-32 & 2.25M \\
VRN-ensemble~\cite{brock2016generative} & 90M & RocNet-1024-32 & 2.53M \\
FusionNet~\cite{hegde2016fusionnet} & 118M & RocNet-2048-32 & 2.80M \\
\hline
\end{tabular}
\end{center}
\caption{Comparison of model size. Numbers of trainable parameters are shown. RocNet is followed by resolution and the leaf size.}
\label{tb:size}
\end{table}

In this section we analyze the model size as well as the computational time and space complexity of our proposed method both analytically and empirically. 

Generally speaking, our method benefits from the recursive structure and hence can be regarded as a lightweight and scalable model. With the increase of input voxel resolution, the number of trainable parameters of our model increases linearly with respect to the logarithm of input/leaf resolutions: $\mathcal{O}(\log (N/k))$ where $N$ is the input resolution and $k$ is the leaf size~(both should be a power of 2). Table~\ref{tb:size} shows the model size of our method~(right column) and other comparison methods~(left column). Since all the other methods were designed specifically for classification, we show the number of trainable parameters of the encoder part of our model for a fair comparison. The total size of our model is approximately twice that shown as the encoder and decoder have roughly the same architecture. For our method we present models for 6 different resolutions. Each method is labeled as RocNet-$N$-$k$ and $k$ is set to $32$. RocNet-256-32, for instance, accommodates an input voxel grid of size $256\times256\times256$. The actual number of trainable parameters shown in Table~\ref{tb:size} coincides with our analysis. Among the comparison methods, VoxNet~\cite{7353481} and LightNet~\cite{zhi2018toward} are also lightweight yet their models only consider a fixed voxel grid of $32\times32\times32$ as they were designed for classification. This resolution should be adequate for a classification task. For more resolution-intensive tasks such as reconstruction and generation, it will lead to severe artifacts and imperfection.

The space and time complexity of our method are $\mathcal{O}((N/k)^3)$ and $\mathcal{O}(\log(N/k))$ 
respectively. The space complexity roughly depends on the number of mixed leaves which, in worst case, is $(N/k)^3$. However, the mixed leaves usually have a very small population in all leaves. Therefore our method enjoys a moderate memory consumption even for $256\times256\times256$ volumes, which easily fits on a modern GPU. Since the nodes within the same depth can be processed in parallel, the computational time for an octree is related to its depth, which is $3\log_8{(N/k)}$.

\section{Experiments and comparisons}

\subsection{Experiments setup}
We evaluate our proposed RocNet by three experiments: 3D shape classification, 3D shape reconstruction and 3D shape generation. The implementation of our algorithm is based on PyTorch~\cite{NEURIPS2019_9015} and TorchFold~\cite{illia_polosukhin_2018_1299387} for dynamic batching. All experiments were done on a server with Intel Core i7-6700K CPU (4.00GHz) and a GeForce GTX Titan X GPU (12GB memory). We set leaf size $k=16$ for voxel size $N=32$ and $k=32$ for all the other input resolutions across all experiments. To clearly demonstrate the advantages of our proposed method, we simply use binary occupancy voxels as input in all experiments. However, the performance can be further improved by employing a more informative input format such as surface normal and truncated signed distance function. Variations of input formats can be easily integrated into our network.

\subsection{3D shape classification}

\begin{table}
\begin{center}
\begin{tabular}{|c|c|c|c|}
\hline
 &  acc. &  &  acc. \\
\hline\hline
3DShapeNets(32) & 77.3\% & RocNet(32-16) & 85.5\% \\
VoxNet(32) & 83.0\%  & RocNet(64-32) & 85.0\% \\
Geometry image & 83.9\% & RocNet(128-32) & 84.4\% \\
OCNN(32) & \textbf{89.6} \% & RocNet(256-32) & 84.1\% \\
\hline
\end{tabular}
\end{center}
\caption{3D shape classification accuracy on ModelNet40 dataset. Voxel-based methods are followed by their tested resolution. RocNet is followed by its resolution and the leaf size.}
\label{tb:classification}
\end{table}

\textbf{Dataset.} We perform the classification task on the ModelNet40 dataset~\cite{wu20153d}. This dataset contains 12,311 labeled CAD mesh from 40 categories. It is split into training~(9843) and testing~(2468) sets and the training set is augmented by rotating each sample 12 times along its upright axis uniformly. We generate voxel representations at four different resolutions for each of the samples: 32, 64, 128, and 256 cubes. %Each voxel model is converted to our defined octree format with a 1D tree topology code and a 4D mixed leaf matrix. 

\textbf{Network architecture.} We simply use the encoder architecture as feature extractor followed by an additional fully-connected layer, a dropout layer and a softmax layer. The input of the fully-connected layer is the output of the tree encoder defined in section~\ref{sec:encoder}.

\textbf{Results and discussion.} The classification results are shown in Table~\ref{tb:classification}. We compare our method with 4 other alternatives: 3DShapeNets~\cite{wu20153d}, VoxNet~\cite{maturana2015voxnet}, Geometry image~\cite{sinha2016deep} and O-CNN~\cite{wang2017cnn}, all of which are voxel-based methods with the exception of Geometry image.

\begin{figure}
\begin{center}
\includegraphics[width=1\linewidth]{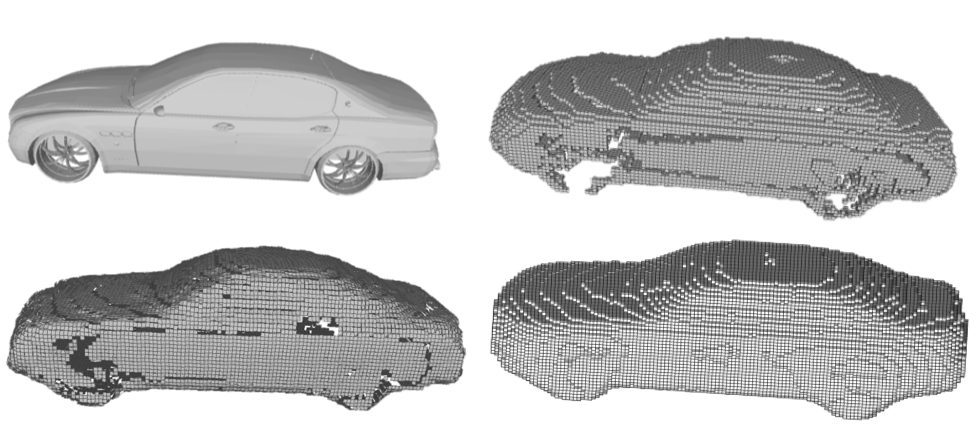}
\end{center}
   \caption[]{Qualitative comparisons. Top: original model, binary O-CNN reconstruction. Bottom: patch-based O-CNN and our result~\footnotemark. $128^3$ is used for all three methods. Our RocNet has fewer missing regions than the other two methods.}
\label{fig:comparison}
\end{figure}

\footnotetext{Pictures of binary O-CNN and patch-based O-CNN are from \cite{wang2018adaptive}.}

%We show the classification accuracy of our proposed network from $32^3$ to $256^3$ in Table~\ref{tb:classification}.
Similar to previous literature, our method achieves the best result when input voxel resolution has size 32 and the performance begins to drop slightly when the size increases. The reason is that categories in ModelNet40 are visually different even at a low resolution, hence less detail is required and $32^3$ is adequate for distinguishing one class from another. Higher resolution probably leads to overfitting.

Our method has inferior accuracy compared to that of O-CNN~\cite{wang2017cnn}, which is partly due to the influence of the input data. In their implementation they used more informative surface normals as input while in our experiments, we simply use the binary occupancy voxel grid. They have shown in their paper that there is approximately a 2 percent drop in accuracy by using binary input. Another reason is that instead of using max-pooling, we employ convolution of stride 2 in the leaf encoder which could lead to fewer patterns being captured. We did an additional experiment where we substituted the 2-stride convolution with 1-stride convolution followed by a max-pooling layer. The accuracy of $32^3$ increases from $85.5\%$ to $86.7\%$. However, this substitution increases the memory requirements of the network. 

Note that focus of this paper is mainly 3D shape autoencoder, we show in this section that by simply using the encoder our method can be easily converted into a lightweight classifier with good accuracy. 

%\subsection{3D shape retrieval}
\subsection{3D shape reconstruction}

\textbf{Datasets.} To evaluate 3D shape reconstruction, we employ 2 datasets: ShapeNet-Car, ShapeNetCorev2~\cite{chang2015shapenet}. ShapeNet-Car contains 7497 car CAD models. ShapeNetCorev2 consists of 39,715 3D models from 13 categories. %ScanNet contains real scanning of 3D environments. We pick 8 environments from ScanNet to demonstrate our method's capability of learning even higher resolutional voxels ($256^3$ and $512^3$).

\textbf{Training protocol.} We use batchsize=50 in training. The average total number of iterations is around 300 for each class. Training takes approximately 20 hours for a class containing 2000 samples in $128^3$ resolution. For both ShapeNet-Car and ShapeNetCorev2 we split training and testing sets by 80\% and 20\%, respectively, as done by \cite{groueix2018papier} and \cite{wang2018adaptive}. Note that the ground-truth node type is used in the training stage to choose the correct decoder. In the test stage, the type of a node is predicted by the trained node classifier.

\begin{figure}
\begin{center}
	\subfloat[$32^3$]{
		\label{fig:car1}
		\centering
		\includegraphics[width=0.33\linewidth]{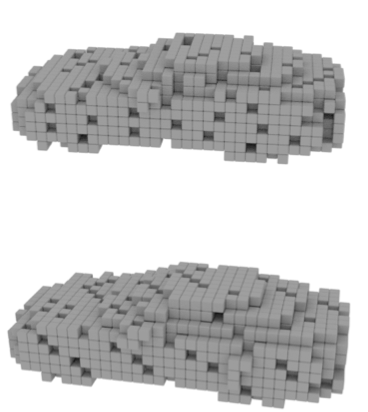}
	}
	\subfloat[$64^3$]{
		\label{fig:car2}
		\centering
		\includegraphics[width=0.33\linewidth]{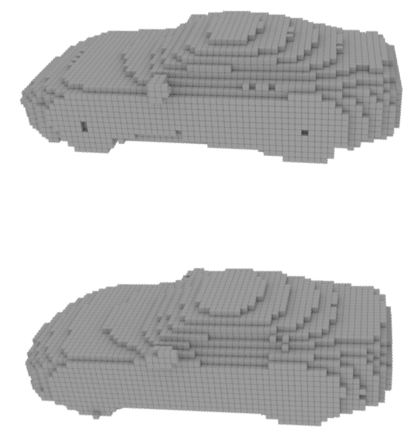}
	}
	\subfloat[$128^3$]{
		\label{fig:car3}
		\centering
		\includegraphics[width=0.33\linewidth]{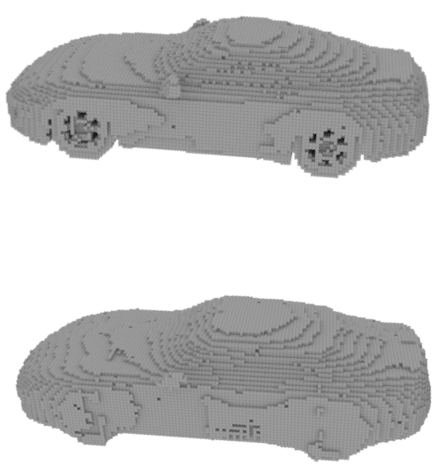}
	}
\end{center}
   \caption{Car reconstruction examples of different resolutions. Top row: ground-truth voxels. Bottom row: reconstructed voxels.} 
\label{fig:car}
\end{figure}

\textbf{Measurements.} We show the qualitative reconstruction results in Figures~\ref{fig:comparison} and \ref{fig:car}. For more visual reconstruction results please refer to the supplementary materials. For quantitative measurements, we use intersection over union (IoU) for ShapeNet-Car and Chamfer distance for ShapeNetCorev2 dataset. This is mainly for the purpose of comparison with existing methods. Using IoU is straightforward since our network produces binary voxels. Chamfer distance is usually used for evaluating the performance of surface-generating methods such as \cite{groueix2018papier,wang2018adaptive}. To compare with this group of methods, we densely sample a set of points, $\mathcal{P}$, from the boundary voxels and a set of points, $\mathcal{G}$, from the ground-truth mesh. The Chamfer distance is calculated as:
\begin{equation}
d(\mathcal{P},\mathcal{G}) = \frac{1}{|\mathcal{P}|} \sum_{x \in \mathcal{P}}{\min_{y \in \mathcal{G}}\| x- y \|^2} + \frac{1}{|\mathcal{G}|} \sum_{x \in \mathcal{G}}{\min_{y \in \mathcal{P}}\| x- y \|^2}.
\end{equation}

\textbf{Results and comparisons.} Figure~\ref{fig:comparison} shows that RocNet produces fewer missing regions~(the wheels) compared to O-CNN and patch-based O-CNN, which explains why RocNet outperforms the rest in the following quantitative analysis. We also observe the reconstructed shapes are more blurred than the ground-truth voxels~(see the details of wheels at $128^3$ resolution in Figure~\ref{fig:car}).

IoU accuracy of the ShapeNet-Car dataset and the that of OGN~\cite{tatarchenko2017octree} is shown in Table~\ref{tb:recons1} and computational requirements are shown in Table~\ref{tb:efficiency}. The Chamfer distance results of ShapeNetCorev2 are shown in Table~\ref{tb:recons2}. 

Table~\ref{tb:recons1} and \ref{tb:efficiency} shows that RocNet outperforms OGN at all three resolutions with lower memory consumption and faster execution time. 
%Examples of reconstruction from different resolutions are shown in Figure~\ref{fig:car}.
Similar to OGN, we also test our algorithm in two modes: with and without octree structure known. In the octree-prediction mode~(named as ``RocNet-p''), the type of each node is predicted by the node classifier while in octree-known mode~(named as ``RocNet-k''), the octree structure is given during decoding. Specifically, the node classifier is disabled in octree-known mode. Intuitively, less information is encoded in octree-known mode hence higher reconstruction accuracy should be achieved. However, as shown in the last two column of Table~\ref{tb:recons1}, these two modes have very close accuracy. This implies that the capacity of the hidden representation is sufficient to store both the octree topology and the leaves. This validates our choice of a $64\times4\times4\times4$ voxels for the hidden representation since using a higher resolution would not lead to a significant increase in accuracy. The rest of our experiments use prediction mode only.

\begin{table}
\begin{center}
\begin{tabular}{|c|c|c|c|c|c|}
\hline
 & OGN-p & OGN-k & Dense & RocNet-p & RocNet-k\\
\hline\hline
32 & 92.4 & 93.9 & 92.4 & \textbf{95.1} & 94.6 \\
64 & 88.4 & 90.4 & 89.0 & \textbf{92.0} & \textbf{92.0} \\
128 & - & - & - & 87.0 & \textbf{87.5} \\
\hline
\end{tabular}
\end{center}
\caption{Reconstruction accuracy on ShapeNet-Car dataset. Numbers shown are intersection over union (IoU\%) between reconstructed and ground-truth voxel in 3 different resolutions. Accuracy for OGN and dense in 128 is not reported in their paper. Boldfaced numbers emphasize the best results.}
\label{tb:recons1}
\end{table}

\begin{figure*}
\begin{center}
\includegraphics[width=1\linewidth]{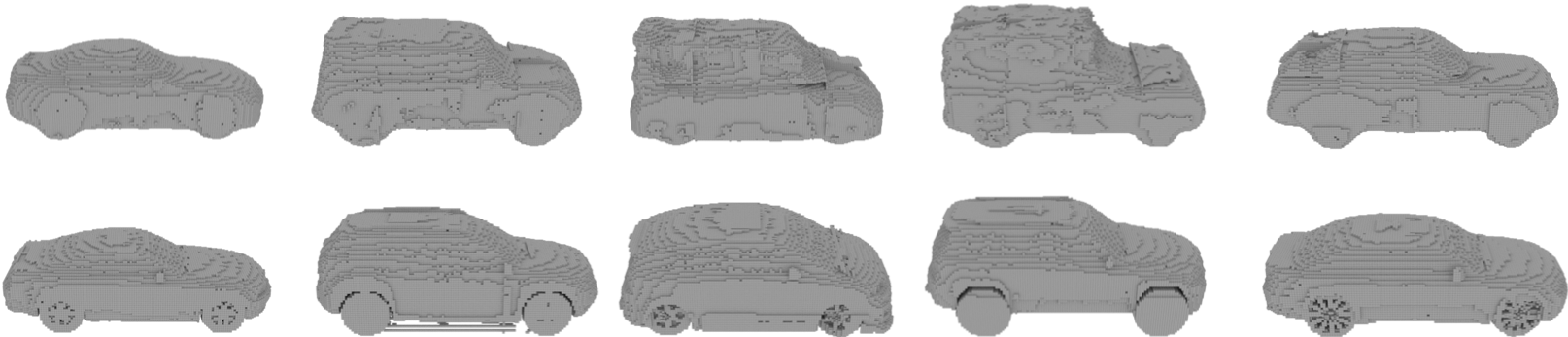}
\end{center}
  \caption[]{Model generation results. Top row: 5 generated models. Bottom row: nearest model in training samples.}
\label{fig:generation}
\end{figure*}

We also report the GPU memory usage and the computational time for one single training iteration in Table \ref{tb:efficiency}. It coincides with the complexity analysis in section~\ref{sec:complexity}. For all three resolutions tested, our algorithm consumes far less memory than OGN. We also observe our algorithm has a larger relative increase compared to OGN. According to the complexity analysis, the memory consumption increases cubically with $N$. We expect our algorithm to take more memory with larger resolutions. However, it is still tractable when processing a $512^3$ voxel grid in our experiments.

\begin{table}
\begin{center}
\begin{tabular}{|c|c|c|c|c|}
\hline
& \multicolumn{2}{c}{Memory(GB)} & \multicolumn{2}{c|}{Time(s)}\\
\hline
 & OGN & RocNet & OGN & RocNet\\
\hline\hline
32	& 0.29 & \textbf{0.017} & \textbf{0.016} & 0.05 \\
64	& 0.36 & \textbf{0.026} & \textbf{0.06}  & \textbf{0.06} \\
128	& 0.43 & \textbf{0.089} & 0.18  & \textbf{0.10} \\
%256	& 0.54 & & 0.64   & \\
%512	& 0.88 & & 2.06   & \\
\hline
\end{tabular}
\end{center}
\caption{Computational efficiency on ShapeNet-Car dataset. Averaged GPU memory usage and time for one iteration are shown. Batch size is set to 1. Boldfaced numbers emphasize the best results.}
\label{tb:efficiency}
\end{table}

Table~\ref{tb:recons2} shows the reconstruction accuracy measured by the Chamfer distance on ShapeNetCorev2 dataset. Compared to the ShapeNet-Car dataset, its samples are more diversified. We compare our method with 4 alternative schemes: PSG~\cite{fan2017point}, AtlasNet~\cite{groueix2018papier} with 125 predicted mesh patches, O-CNN~\cite{wang2017cnn} and Adaptive O-CNN~\cite{wang2018adaptive}, among which PSG is the only point set generating method while the others generate meshes. For comparison, we calculate the Chamfer distance in the same protocol as for O-CNN. RocNet has the best performance in most categories followed by Adaptive O-CNN and AtlasNet. It is worth noticing that since RocNet generates voxels instead of a mesh, there is an intrinsic inaccuracy between the cube-like voxels and the smooth mesh. Mesh-based methods such as Adaptive O-CNN and AtlasNet do not suffer from this drawback. We expect our method to perform even better with a patch-based representation, which is an area for future work.

\begin{table*}
\begin{center}
\begin{tabular}{|l|c|c|c|c|c|c|c|c|c|c|c|c|c|c|}
\hline
 & avg. & pla. & ben. & cab. & car. & cha. & mon. & lam. & spe. & fir. & cou. & tab. & cel. & wat. \\
\hline\hline
PSG	& 1.91	& 1.11 & 1.46 & 1.91 & 1.59 & 1.90 & 2.20 & 3.59 & 3.07 & 0.94 & 1.83 & 1.83 & 1.71 & 1.69 \\
AtlasNet(125) & 1.51 & 0.86 & \textbf{1.15} & 1.76 & 1.56 & 1.55 & 1.69 & \textbf{2.26} & 2.55 & \textbf{0.59} & 1.69 & 1.47 & 1.31 & 1.23 \\
OCNN & 1.60 & 1.12 & 1.30 & 1.06 & 1.02 & 1.79 & 1.62 & 3.71 & 2.56 & 0.98 & 1.17 & 1.67 & 0.79 & 1.88\\
Adaptive OCNN & 1.44 & 1.19 & 1.27 & \textbf{1.01} & 0.96 & 1.65 & 1.41 & 2.83 & 1.97 & 1.06 & 1.14 & 1.46 & 0.73 &	1.82\\
\hline
RocNet & \textbf{1.05} & \textbf{0.49} & 1.34 & 1.09 & \textbf{0.83} & \textbf{1.29} & \textbf{0.74} & 2.32 & \textbf{1.58} & 0.73 & \textbf{0.80} & \textbf{0.91} & \textbf{0.72} & \textbf{0.82} \\		
\hline
\end{tabular}
\end{center}
\caption{Chamfer distance tested on ShapeNetCorev2 dataset. The Chamfer distance is multiplied by $10^3$ for better display. Boldfaced numbers emphasize the best results. Results of PSG and AtlasNet are from \cite{groueix2018papier}. Results of O-CNN and Adaptive-OCNN are provided in \cite{wang2018adaptive}.}
\label{tb:recons2}
\end{table*}

\subsection{3D shape generation}

In this section we present the results of model generation using RocNet. Figure~\ref{fig:generation} shows 5 generated models by RocNet autoencoder trained on ShapeNet-Car dataset. 

We generate new models by randomly sampling within the convex hull of the trained samples in 80D feature space. Please note that the octree topology is generated on the fly in this experiment since it is impossible to obtain the ground-truth which exists in reconstruction task. From Figure~\ref{fig:generation} we could observe that RocNet is able to establish a semantically plausible shape manifold where each generated model can be visually identified as a car. We also present the corresponding training model closest to the sampled point in feature space. Some generated models are very similar to existing models such as the first example. There also exist some examples where there are visible differences between existing ones. This indicates that RocNet has captured some semantic information from training samples. 

We also notice that the generated models are more blurred than existing ones such as the absence of the details of the wheels. This is a general problem an autoencoder suffers from without an adversarial~(GAN~\cite{goodfellow2014generative}) term. Currently our RocNet has the structure of a standard autoencoder. It would be possible to combine our current loss function with an adversarial loss in the future.

\section{Conclusion and future work}
We propose a recursive octree-based network which encodes and decodes octrees by either recursively merging or producing octants. Our key insight is the recursive nature of the proposed network fits the hierarchical feature of octrees well. By using an octree representation we are able to save a large amount of storage and computational time and higher accuracy is achieved compared to existing methods for 3D shape reconstruction and generation. The architecture also provides good results for 3D shape classification. We demonstrated the advantages of our method by three experiments. The complexity of our method is analyzed both theoretically and empirically.

For future work we would like to make variants of RocNet by using more advanced network structures such as variational autoencoder~(VAE) and generative adversarial networks~(GAN). We believe by adding a GAN term RocNet will be able to generate more realistic models with more details. RocNet then can be applied to more realistic 3D content generation tasks.

Another promising improvement to RocNet is to use more input information. As suggested by existing work~\cite{wang2019normalnet,wang2017cnn}, using more informative input information such as normals or a truncated signed distance field leads to better performance. RocNet is compatible with arbitrary input formats as long as it dose not violate the sparseness of 3D data. This sparseness arises naturally from the fact that surfaces in the world form manifolds in 3D space. We will also explore the possibility of directly generating meshes as output of the network.

%------------------------------------------------------------------------

\bibliographystyle{unsrt}  
\bibliography{references}

\end{document}